\documentclass{IEEEcsmag}

\pdfoutput=1

\usepackage[colorlinks,urlcolor=blue,linkcolor=blue,citecolor=blue]{hyperref}
\expandafter\def\expandafter\UrlBreaks\expandafter{\UrlBreaks\do\/\do\*\do\-\do\~\do\'\do\"\do\-}
\usepackage{upmath,color}
\usepackage{balance}

\jvol{XX}
\jnum{XX}
\paper{}
\jmonth{Oct}
\jname{Publication Name}
\jtitle{Integrating Graphs with Large Language Models: Methods and Prospects}
\pubyear{2023}

\begin{document}


\title{Integrating Graphs with Large Language Models: Methods and Prospects
}

\author{Shirui Pan}
\affil{Griffith University, Gold Coast, QLD 4215, Australia}

\author{Yizhen Zheng \& Yixin Liu}
\affil{Monash University, Melbourne, VIC 3800, Australia}


\markboth{THEME/FEATURE/DEPARTMENT}{THEME/FEATURE/DEPARTMENT}

\begin{abstract}
Large language models (LLMs) such as GPT-4 have emerged as frontrunners, showcasing unparalleled prowess in diverse applications, including answering queries, code generation, and more. Parallelly, graph-structured data, an intrinsic data type, is pervasive in real-world scenarios. Merging the capabilities of LLMs with graph-structured data has been a topic of keen interest. This paper bifurcates such integrations into two predominant categories. The first leverages LLMs for graph learning, where LLMs can not only augment existing graph algorithms but also stand as prediction models for various graph tasks. Conversely, the second category underscores the pivotal role of graphs in advancing LLMs. Mirroring human cognition, we solve complex tasks by adopting graphs in either reasoning or collaboration. Integrating with such structures can significantly boost the performance of LLMs in various complicated tasks. We also discuss and propose open questions for integrating LLMs with graph-structured data for the future direction of the field.
\end{abstract} 

\maketitle

\begin{figure*}
    \centering
    \includegraphics[width=1.0\linewidth]{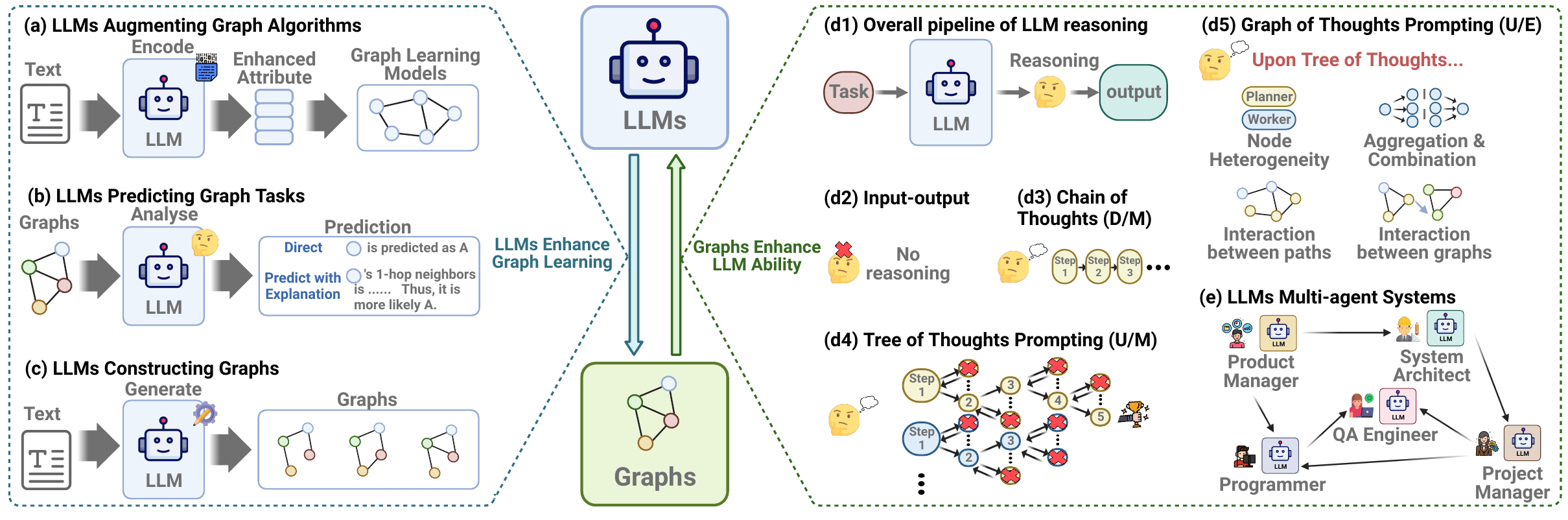}
    \caption{The overall framework of the mutual enhancement between LLMs and Graphs. (a)-(c): three pathways for LLMs to enhance graph learning. (d)-(e): techniques for graph structures enhancing LLM reasoning. Brackets after technique names indicate graph types. D, U, M and E represent directed, undirected, homogeneous and heterogeneous graphs, respectively.}
    \label{fig:overall}
\end{figure*}

\chapteri{L}arge language models (LLMs) have rapidly taken centre stage due to their remarkable capabilities. They have demonstrated prowess in various tasks, including but not limited to, translation, question-answering, and code generation. Their adaptability and efficiency in processing and understanding vast amounts of data position them as revolutionary tools in the age of information. Concurrently, graphs are a natural representation of the world, and their ability to capture complex relationships between entities makes them a powerful tool for modelling real-world scenarios.  For instance, structures reminiscent of graphs can be observed in nature and on the internet. Given the individual significance of both LLMs and graph structures, the exploration into how they can be synergistically combined has emerged as a hot topic in the AI community. 

In the paper, we delineate two primary paradigms for the synergy between LLMs and graphs. The first paradigm, ``LLMs enhance graph learning'' involves harnessing the capabilities of LLMs to handle various graph-related tasks.  This includes predicting graph properties, such as the degrees and connectivity of nodes, as well as more intricate challenges like node and graph classification.  Here, LLMs can either supplement graph algorithms or serve as main predictive/generative models. Conversely, the second paradigm, ``Graphs advance LLMs ability'', capitalises on the inherent structure of graphs to enhance the reasoning capabilities of LLMs or help LLMs to collaborate, facilitating them in handling multifaceted tasks. By leveraging graph structures, the efficacy of LLMs in complex problem-solving can be significantly augmented.

\subsection{Why integrating graphs and LLMs?}

\textbf{Using LLMs to address graph tasks} enjoys two primary advantages. 
First, unlike the often opaque graph deep learning techniques, LLMs approach graph-related challenges primarily through reasoning, providing clearer insight into the basis for their predictions. This transparency offers a more interpretable method for understanding complex graph analyses. 
Secondly, LLMs possess an extensive repository of prior knowledge spanning diverse domains. Traditional graph learning models, constrained by limited training data, struggle to comprehensively assimilate this wealth of knowledge. Consequently, harnessing LLMs for graph data processing presents an opportunity to leverage their scalability and the expansive reservoir of prior knowledge. 
Such knowledge can be especially valuable for graph machine learning in domains such as finance and biology.

\noindent \textbf{Using graphs to enhance LLMs} is also a promising learning paradigm. 
In specific, graphs can substantially amplify the capacity of LLMs in both logical reasoning and collaboration within multi-agent systems$^{1,2,7,9}$. 
For instance, using a straightforward prompt like "Let's think step by step", commonly referred to as the chain-of-thoughts, has been proven to markedly enhance the LLM's proficiency in resolving mathematical problems$^4$. It's noteworthy that such enhancements are observed even with the use of a chain, which represents one of the simplest graph structures. This gives rise to the anticipation that leveraging more intricate graph structures could usher in even more profound improvements. From a broader viewpoint, in multi-agent systems, graphs model inter-agent relationships, facilitating efficient information flow and collaboration.

\section{LLMs Enhance Graph Learning}
One pivotal approach to integrating LLMs and graphs involves leveraging LLMs to bolster graph learning. As illustrated in the left part of Figure~\ref{fig:overall}, this enhancement can materialise through three distinct pathways: augmenting conventional graph algorithms with the prowess of LLMs; directly employing LLMs for downstream graph-related tasks; and utilizing LLMs in the intricate construction of graph structures. In the following sections, we dissect each of these strategies in detail.


\subsection{LLMs Augmenting Graph Algorithms}
The integration of Large Language Models (LLMs) with graph algorithms primarily seeks to harness LLMs as attribute-enhancement mechanisms, elevating the intrinsic attributes of graph nodes. As depicted in Figure~\ref{fig:overall}(a), LLMs process text information for nodes to produce refined attributes. These enhanced attributes can potentially improve the performance of graph learning models such as graph neural networks (GNNs).

A direct approach is to employ LLMs as encoders for processing node text-based attributes, with the option to fine-tune on specific downstream tasks$^3$. Another technique uses a proprietary LLM, GPT-3.5, to simultaneously produce predictions and explanations for tasks like paper classification$^5$. Using another open-source LLM, they derive node embeddings by encoding both the output of LLMs and the original attributes. These embeddings are combined and then integrated into GNNs to boost performance. 

A more sophisticated approach uses an iterative method to harmoniously integrate both GNNs and LLMs capabilities$^3$. They are initially trained separately; then, via a variational EM framework, the LLM uses text and GNN's pseudo labels, while the GNN utilizes LLM-encoded embeddings or node attributes and LLMs' pseudo labels, iteratively boosting mutual performance.


\subsection{LLMs Predicting Graph Tasks}
LLMs are adept at predicting graph properties, including attributes like node degrees and connectivity, and can even tackle complex challenges such as node and graph classification, as illustrated in Figure~\ref{fig:overall}(b).

A straightforward application involves presenting LLMs with zero-shot or few-shot prompts, prompting them to either directly predict an outcome or to first provide an analytical rationale followed by the ultimate prediction$^{3,6}$. Experiments reveal that while LLMs demonstrate a foundational grasp of graph structures, their performance lags behind that of graph neural network benchmarks. They also show that performance of LLMs is significantly affected by the prompting strategy and the use of graph description language, which is a textual way to describe graphs.

A more advanced method, dubbed InstructGLM, has been put forth$^8$. This strategy utilises a multi-task, multi-prompt instructional tuning process to refine LLMs prior to inference on specific tasks. During fine-tuning, nodes are treated as new tokens—initialised with inherent node features—to broaden the original vocabulary of LLMs. Consequently, node embeddings can be refined during the training phase. Employing this refined methodology, their system outperforms graph neural network benchmarks across three citation networks.

\subsection{LLMs Constructing Graphs}
LLMs can help in building graphs for downstream tasks as shown in Figure~\ref{fig:overall}(c). For instance, some researchers have tried using LLMs to analyse news headlines and identify companies that might be impacted$^{10}$. In specific, a network of companies that have correlations is constructed by LLMs automatically. The generated network can be used to improve the performance of predictions of stock market movements.

\section{Graphs Enhance LLM Ability}
Leveraging graph structures can significantly boost the reasoning and collaborative capacities of LLMs. As shown in the right part of Figure~\ref{fig:overall}, these improvements emerge via two primary mechanisms: (1) employing graph structures to bolster logical reasoning in LLMs, and (2) utilizing graph structures to enhance LLM collaboration in multi-agent systems. We delve deeper into each of these approaches in the subsequent sections.


\subsection{Graphs Improving LLMs Reasoning}
Graphs are the foundational structure of human reasoning. Through tools like mind maps and flowcharts, and strategies like trial and error or task decomposition, we manifest our intrinsic graph-structured thought processes. Not surprisingly, when properly leveraged, they can significantly elevate the reasoning capabilities of LLMs. As illustrated in Figure~\ref{fig:overall}(d1), when tasked, LLMs follow a sequence: they process the input data, engage in reasoning, and then produce the final results. Figure~\ref{fig:overall}(d2) highlights the limitations of LLMs using ``Input-output Prompting''; without reasoning, their performance tends to suffer, especially with complex tasks.

Employing graph structures, from basic chains and trees to more complex designs, can profoundly augment the reasoning capabilities of LLMs. Consider the ``chain-of-thought prompting'' (COT) method, depicted in Figure~\ref{fig:overall}(d3)$^4$. In this, LLMs harness a chain, a type of directed acyclic graph, for structured problem-solving. Remarkably, even this basic framework triples LLMs' efficacy on GSM8K, a math word problem benchmark.

In contrast, the ``Tree of Thoughts'' (ToT) method, utilising trees—an elementary undirected acyclic graph—delves deeper into reasoning. Eeach reasoning phase in ToT is a node$^7$. LLMs traverse this tree, eliminating non-compliant nodes and returning upwards as necessary, to deduce the solution. With this methodology, LLMs notch up a 74\% accuracy in the ``Game of 24'' test, overshadowing the 4\% from COT$^7$.

Diving into intricate graph structures propels LLMs' capabilities even further. Improving ToT, the ``Graph of Thoughts'' (GoT) paradigm has been introduced$^{1,2}$, as illustrated in Figure~\ref{fig:overall}(d5). This advanced reasoning graph can be heterogeneous, with diverse nodes dedicated to specific tasks. Sophisticated mechanisms, such as node aggregation and combination (A\&C), and dynamic interactions between paths and graphs, are incorporated. A\&C, for instance, facilitates node subdivision for task decomposition and node amalgamation$^1$. Path interactions offer LLMs greater flexibility by enabling cross-path traversals, a leap from ToT's isolated branch framework. Multi-graph interactions can even be orchestrated for intricate tasks$^2$. These GoT methodologies dramatically outpace simpler graph models in handling complex challenges, indicating that more intricate graph structures could usher in even more significant enhancements.

\subsection{Graphs Building LLMs Collaboration}
While the preceding section examined the capabilities of individual LLMs, complex tasks, such as software development, require multiple LLMs to work in tandem within a collaborative framework, i.e., multi-agent systems, as illustrated in Figure~\ref{fig:overall}(e). Graph structures can be instrumental in this context. As depicted in the same figure, these structures can effectively model the relationships and information flow between collaborating LLMs.

\section{Open Questions and Directions}
The intersection of Large Language Models (LLMs) with graph structures holds promise, yet its current development sparks some open questions and challenges. 

\subsection{LLMs Enhancing Graph Learning}
\subsubsection{\textbf{Question~1.}} \textbf{How to leverage LLMs to learn on other types of graphs beyond Text-attributed Graphs (TAG)?} Current LLMs for graph learning primarily concern TAGs. However, real-world graph data, such as social networks and molecular graphs, often incorporate attributes from different domains. To realise the potential of LLMs in graph learning, it is crucial to efficiently handle a wide variety of graphs as input to these models. 

\noindent\textbf{Future Directions:} \textbf{Direction~1.} Translate diverse data types into textual format: For instance, a user's profile on a social network might list attributes like age, address, gender, and hobbies. These can be articulated as: "User X is a male in his 20s, residing in Melbourne, with a passion for playing guitars." \textbf{Direction~2.} Leveraging multi-modal models for graph-text alignment: Multi-modal LLMs have already made notable strides in domains like audio and images. Identifying methods to synchronise graph data with text would empower us to tap into the capabilities of multi-modal LLMs for graph-based learning.

\subsubsection{\textbf{Question~2.}} \textbf{How can we help LLMs understand graphs? }Central to the success of LLMs in graph learning is their ability to genuinely comprehend graphs. Experimental evidence suggests that the choice of graph description language can have a significant impact on LLM performance$^6$.

\noindent\textbf{Future Directions:} \textbf{Direction~1.} Expanding graph description languages: Current graph description languages offer a somewhat restricted scope. Developing enhanced description methods would enable LLMs to grasp and process graphs more effectively. \textbf{Direction~2.} Pretraining or fine-tuning LLMs on Graphs: Pretraining or fine-tuning LLMs on various graph data converted by graph description language can help LLMs understand graphs better$^8$. \textbf{Direction~3.} Foundational graph models for graph learning: While foundational models have made strides in areas like language, and image, a gap remains in establishing large-scale foundational models for graphs. Converting graphs to textual format offers a unique opportunity: LLMs can be trained on this data, enabling graph learning to leverage the prior knowledge and scalability inherent in LLMs.

\subsection{Graphs Enhance LLM Ability}
\subsubsection{\textbf{Question~3.}}  \textbf{How to elevate more sophisticated graph structures to enhance LLM reasoning?} Current explorations into LLM reasoning have touched upon graph structures like chains, trees, and traditional graphs. However, there is vast potential in delving into more intricate graph structures, such as hypergraphs, probabilistic graphical models, and signed graphs. 

\noindent\textbf{Future Directions:} Expanding the types of graphs for LLM reasoning: Diversifying the graph types used could significantly bolster LLM reasoning. 

\subsubsection{\textbf{Question~4}.} \textbf{How to elevate more sophisticated graph structures to enhance multi-agent systems (MLS)?} Presently, the graph structures guiding MLS, like that in MetaGPT$^9$, are relatively rudimentary. While MetaGPT employs the waterfall model in software development—illustrated by a simple chain structure linking different agents—contemporary software development is far more nuanced with intricate agent relationships and multifaceted processes.

\noindent\textbf{Future Directions:}  Incorporating advanced graph structures for team-based LLM workflows: Drawing from the utility of graph structures in reasoning, adopting varied graph forms such as trees, traditional graphs, and even more intricate structures may help.

\subsubsection{\textbf{Question~5.}} \textbf{How to integrate graph structures into the pipeline of LLMs?} The applicability of graph structures is not confined to reasoning and collaboration. There's a compelling case to be made for their integration across all stages of the LLM lifecycle: from training and fine-tuning to inference.

\noindent\textbf{Future Directions:}  Utilising graph structures in training, fine-tuning, and inference. For example, graphs can structure the training data, enabling more effective learning.

\def\refname{REFERENCES}

\vspace*{-8pt}
\balance




\end{document}